\documentclass{ecai}

\usepackage{latexsym}
\usepackage{amssymb}
\usepackage{amsmath}
\usepackage{amsthm}
\usepackage{booktabs}
\usepackage{enumitem}
\usepackage{graphicx}
\usepackage{color}

\usepackage{multirow}

\newcommand{\BibTeX}{B\kern-.05em{\sc i\kern-.025em b}\kern-.08em\TeX}

\begin{document}


\begin{frontmatter}


\paperid{2145} 


\title{The Propensity for Density in Feed-forward Models}


\author[A,B]{\fnms{Nandi}~\snm{Schoots}
\thanks{Corresponding Author. Email: \url{nandischoots@gmail.com}}\footnote{Equal contribution.}}
\author[A,B]{\fnms{Alex}~\snm{Jackson}
\footnotemark}
\author[C]{\fnms{Ali}~\snm{Kholmovaia}}
\author[A]{\fnms{Peter}~\snm{McBurney}
}
\author[B]{\fnms{Murray}~\snm{Shanahan}
}

\address[A]{King's College London, United Kingdom}
\address[B]{Imperial College London, United Kingdom}
\address[C]{Philipps-Universit\"at Marburg, Germany}


\begin{abstract}
    Does the process of training a neural network to solve a task tend to use all of the available weights even when the task could be solved with fewer weights? 
    To address this question we study the effects of pruning fully connected, convolutional and residual models while varying their widths. 
    We find that the proportion of weights that can be pruned without degrading performance is largely invariant to model size.
    Increasing the width of a model has little effect on the density of the pruned model relative to the increase in absolute size of the pruned network.
    In particular, we find substantial  prunability across a large range of model sizes, where our biggest model is 50 times as wide as our smallest model.
    We explore three hypotheses that could explain these findings.\\Source code: \url{https://github.com/alexjackson1/simplicity}.
\end{abstract}

\end{frontmatter}



\section{Introduction} \label{sec:intro}

Large, dense neural networks are capable of impressive performance \citep{samekExplainingDeep2021}. However, there is both a high computational cost associated with deploying such networks and a lack of transparency about their behaviour \citep{liptonMythosModel2018}.
Accordingly, sparsity has become a target for a variety of researchers aiming to reduce cost and increase the interpretability of deep learning systems.

According to Parkinson's law 
the time taken to complete a bureaucratic job expands to fill the time available to do it.
In this paper, we ask whether an analogous law holds for neural network capacity. 
Does a trained network tend to use all of the parameters available to solve a task even if the task could be solved with fewer parameters?



We investigate how many weights of a trained network $f(x ; \theta)$ can be set to zero before the model substantially loses performance.
In particular, we prune away the lowest magnitude weights until the model's accuracy drops by 5\% (without retraining the remaining subnetwork).
The resulting subnetwork is what we consider the `core' model with parameters $\omega$ and we call the proportion of remaining weights $\frac{|\omega|}{|\theta|}$ the \emph{effective density} of the model.

Given an architecture, training regime and task, our null hypothesis is that the `core' model size $|\omega|$ does not change with the network width. 
That is, the null hypothesis says, for an effective density $\frac{|\omega|}{|\theta|}$, the denominator $|\theta|$ is the only quantity that changes.

We evaluate whether models tend towards density or sparsity by varying the width of a model (between 0.1x and 5x the default width) and assessing how the effective densities change as the model size increases.
Note that the high performing subnetworks for a model architecture of size 0.1x are structurally also subnetworks of size 5x (because the 5x models are simply wider).
In other words,
the solutions for a model architecture of a given size are also valid solutions for wider models that use the same architecture.

We find that as we increase the width, the `core' model size $|\omega|$ vastly increases. 
This contradicts our null hypothesis, which says that the `core' model size remains constant. 
Accordingly, we find that the effective densities for different model sizes are much more similar than they would be if the null hypothesis were true. 
Specifically, we find no 0.1x solutions when we train a 5x model, 
implying that certain solutions are favoured when training an architecture of a particular width.

To investigate this phenomenon, we explore three independent explanations: 
first, that the initialisation locks in a particular density; 
second, that the models implement substantially different functions; 
and 
third, 
that wider models use the abundance of parameters at their disposal to train more \textit{monosemantic} units.
We find that the first and last explanation may play a role for convolutional models trained with Adam, while the second may be relevant for fully connected models trained with SGD.

Our contributions are the following:
\begin{itemize} [topsep=0pt]
    \item In Section \ref{sec:main-finding} we reject the null hypothesis that the `core' model size does not change with the network width for the models we investigate, and report on the pruning trajectories and effective densities.
    \item In Section \ref{sec:hypotheses} we discuss and evaluate three potential explanations for this finding, observing varying results across different architectures and optimizers.
\end{itemize}

\section{Background and Related Work}\label{sec:related-work}



\paragraph{Pruning and Distillation.}

The work most closely related to ours investigates the robustness of networks to random ablation of neurons by randomly pruning them,
rather than performing magnitude-based weight pruning as we do.
\citet{casperFrivolousUnits2021} vary the width of a network between 0.25x and 4x the default width, train the network,
and apply a randomly generated pruning mask: they report what percentage of test labels do not change for different mask sizes.
The results show the effective density decreases as the number of units increases. 
We expand on this line of work by: 1) investigating how the proportion of crucial parameters, as opposed to neurons, changes as the model increases in width; and 2) pruning parameters with a heuristic (magnitude pruning) as opposed to randomly ablating.

In their investigation of the effect of pruning on downstream tasks, \citet{jaiswalEmergenceEssential2023} tangentially show that, for different sizes of OPT models \citep{zhangOPTOpen2022} (125m, 350m, 1.3B), the effective densities of the models are consistent with our findings.
In contrast, we test a specific hypothesis relating density and \emph{width} (as opposed to size); and also conjecture and test possible explanations for this relationship.

Our aims sit in contrast to the typical focus of neural network pruning \citep{blalockWhatState2020}: retaining model capabilities with a smaller compute budget.
Specifically, instead of making networks sparser to, for instance, reduce the storage footprint of the network or the computational cost associated with inference \citep[e.g.][]{jacksonFindingSparse2023},
we prune with the intention of uncovering the model's `true' size or effective density.

The field of knowledge distillation \citep{gouKnowledgeDistillation2021} aims to compress a large teacher model into a small student model by training the student model to mimic the behavior of the teacher model.
The method of compressing a teacher model into an ever smaller student model can be used to approximate the function complexity of the teacher model, but can not capture our focus in this paper: the `size' of the implementation of the teacher model.



\paragraph{Implicit Inductive Bias.}

Given a dataset and network architecture, there is a hypothesis space of  models that can be considered by a learning algorithm when it is tasked with fitting this data. 
Any training regime inherently has \emph{implicit inductive biases} that make it more likely one model is selected from the hypothesis space than another.
We call a hypothesis that is selected by a training process a \emph{solution}.


The parameter-mapping function \citep{perezDeepLearning2019}
and initialization scheme \citep[e.g.][]{heDelvingDeep2015}
are important forms of inductive bias in deep learning.
Examples of recent insights about 
implicit regularization include the uncovering of grokking \citep{powerGrokkingGeneralization2022} and double descent \citep{nakkiranDeepDouble2020}.
Based on our findings we hypothesize that models have an implicit inductive bias towards using up a large portion of the available parameters.

\paragraph{Strong Lottery Ticket Hypothesis.}



The Strong Lottery Ticket Hypothesis states that: 
`Within a sufficiently overparameterized neural network with random weights (e.g. at initialization), 
there exists a subnetwork that achieves competitive accuracy' \citep{ramanujanWhatHidden2020}.
This conjecture has been proven under certain assumptions \citep{malachProvingLottery2020}.

\citet{ramanujanWhatHidden2020} empirically find that subnetworks of randomly initialized Conv-2 networks can solve CIFAR-10 to a decent accuracy level.
They introduce an algorithm they call `edge-popup' to find well-performing subnetworks. 
The best subnetworks they find have a size of around 50\% of the total network parameters, but they find well-performing subnetworks in the region of 30\% to 70\%. 
They hypothesize this is because there are more subnetworks of size 50\% than of any other size, because of combinatorics.
They use five random seeds and report on the mean and standard deviation.
Note that their algorithm does not employ a brute-force method to search for subnetworks; rather, it utilizes a sophisticated approach for learning an effective mask, which may have its own inductive biases.

Suppose that instead of training a model (which can involve an inductive bias)
we randomly initialized a network and sampled subnetworks 
until we found a subnetwork with low loss.
What would then be the distribution of `winning' subnetwork sizes?
The result in \citet{ramanujanWhatHidden2020} suggests that without any inductive bias (from a learning algorithm) and by just brute-force sampling, we may find that the winning networks sizes would be distributed around 50\%.
Unfortunately, it is not possible to conclude this since only the best subnetworks are reported on (existence) rather than the frequency by which they find good subnetworks of a given size (abundance).

Summarizing, the above suggests that when we increase the width of a network, the `core' model size may increase proportionally, which is opposite to our null hypothesis.



\section{Experimental Setting}\label{sec:experimental-setting}

For the results presented in this work, we train a variety of small models for classification tasks  \citep{kingscollegelondonKingComputational2022}, after which we prune them in a stepwise manner and record the pruned model's performance.
Note that, in contrast to the literature \citep{frankleLotteryTicket2019}, during magnitude pruning we do \emph{not} train or finetune in between pruning steps.
This is because we want to capture the `size' of the implementation that was learned as opposed to e.g. finding the smallest possible network that can implement the same function.

\subsection{Optimization Problem}
We explore fully connected, convolutional and residual network architectures, and use three optimizers and two initialization schemes across our experiments.
As discussed, we also vary the width of these architectures from their default values using a scaling factor.

\paragraph{Datasets.}
We use the MNIST handwriting recognition dataset \citep{dengMNISTDatabase2012} and 
CIFAR-10 image dataset \citep{krizhevskyLearningMultiple2009} as the tasks to be optimized.
MNIST and CIFAR-10 respectively contain 
70,000 and 60,000 labelled images that includes a test set of 10,000 samples each.
For both datasets, we randomly partition a 5,000 image validation set from the remaining training set.

\paragraph{Default Architectures.}\label{sec:default-architectures} 
We train fully-connected dense networks, as well as convolutional and residual networks. 
Below, we describe the \emph{default} architectures we use.
The scaling factor is applied to the widths mentioned in these default architectures.

\begin{itemize}
    \item The fully-connected dense networks are based on a LeNet-300-100 architecture \citep{lecunGradientbasedLearning1998} and consist of two dense layers with 300 and 100 units respectively (with ReLU activations). We refer to these networks as multi-layer perceptrons (MLPs).
    \item The convolutional networks are based on a variation \citep{frankleLotteryTicket2019} of the VGG architecture 
    \citep{simonyanVeryDeep2015}
    referred to as Conv-2.
    These networks consist of two convolutional layers with 64 filters, a max-pooling layer, followed by two dense layers with 256 units and ReLU activations. 
    \item The residual networks are based on the ResNet-18 architecture used in \citep{frankleLotteryTicket2019}. The first layer is a convolutional layer, followed by a batch norm layer and ReLU activation; there are then three residual blocks with two convolutional layers (again each followed by a batch norm layer and ReLU activations) and an average pool layer; finally, a dense layer projects the flattened output into 10 classes.
    By default, all convolutional layers have 16 filters.
\end{itemize}

The above default widths are taken from the literature but ultimately serve as an arbitrary reference point to compare the effect of scaling the layer widths.

\begin{table}[t]
\caption{The absolute number of parameters between the first layer and output layer as a result of increasing the number of units in each layer. The 5x MLP model is roughly 80.7 times as large as the 0.1x model.}
\vspace{0.5cm}
\centering
\resizebox{0.7\columnwidth}{!}{
    \begin{tabular}{|c||r|r|r|}
        \hline
        \textbf{Size (x)} & \multicolumn{1}{c|}{\textbf{MLP}} & \multicolumn{1}{c|}{\textbf{Conv-2}} & \multicolumn{1}{c|}{\textbf{ResNet}} \\ \hline \hline
        0.1 & 23,920    & 536,963     & 27,762  \\ \hline
        0.5 & 125,600   & 11,388,288  & 221,586 \\ \hline
        1.0 & 266,200   & 43,780,864  & 443,122 \\ \hline
        2.0 & 592,400   & 171,578,880 & 886,194 \\ \hline
        5.0 & 1,931,000 & N.A.        & N.A.    \\ \hline
    \end{tabular}
}
\label{tab:number-of-parameters}
\end{table}

\begin{figure*}[t]
    \centering
    \includegraphics[width=0.98\linewidth]{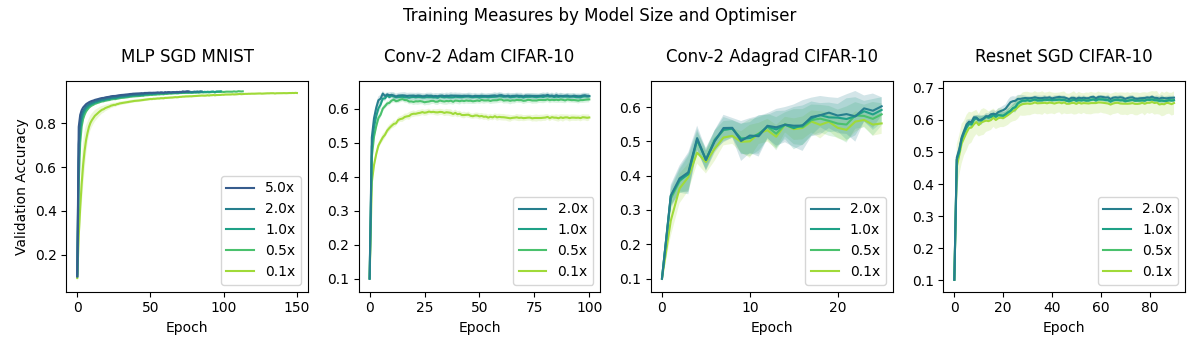}
    \caption{The x-axis shows the epochs and the y-axis shows the validation accuracy. We plot data for models of 5 different architecture sizes (0.1, 0.5, 1, 2, 5).}
    \label{fig:main-training-acc}
\end{figure*}

\begin{figure*}[t]
    \centering
    \includegraphics[width=0.98\linewidth]{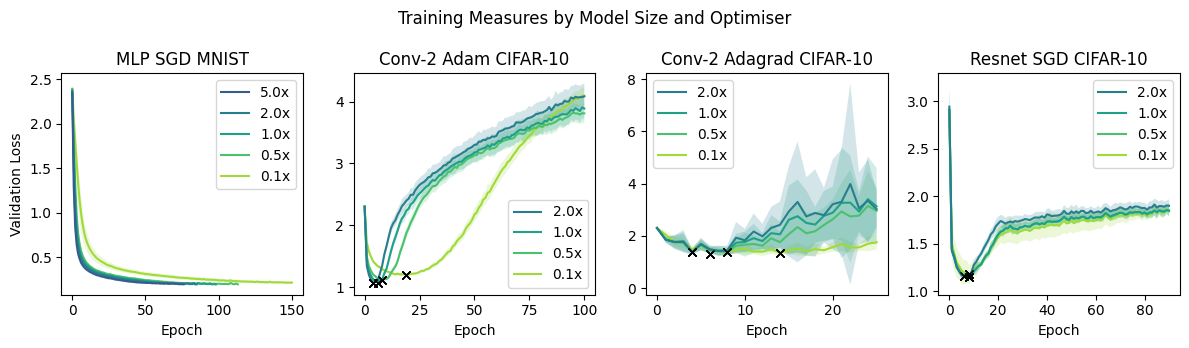}
    \caption{The x-axis shows the epochs and the y-axis shows the validation loss. We plot data for models of 5 different architecture sizes (0.1, 0.5, 1, 2, 5).}
    \label{fig:main-training-loss}
    \vspace{0.5cm}
\end{figure*}
\vspace{0.5cm}

\newpage

\vspace{1cm}
\paragraph{Network Size.} 
The width of the layers at initialization is determined by multiplying the default number of units in each layer by a scaling factor named \emph{size}. 
Mechanically, for the dense layers, this means multiplying the layer's width by the size; for the convolutional layers, this means multiplying the number of filters by the size.
In this paper, we consider five sizes of 0.1, 0.5, 1, 2 and 5.
For the convolutional and residual networks, we were unable to train models of size 5 on a single NVIDIA A100 GPU and these results are therefore absent.
Table \ref{tab:number-of-parameters} shows the number of parameters for each size of network.

\paragraph{Initialization Schemes.} 
We use both 
Glorot \citep{glorotUnderstandingDifficulty2010} and 
He \citep{heDelvingDeep2015} initialization schemes. 
To save resources we only used Glorot for ResNets.

\paragraph{Optimizers.} 
We train each model using one of three different gradient descent optimizers:
\begin{itemize}
    \item Stochastic gradient descent (SGD) with a learning rate of $1\times 10^{-3}$ for solving MNIST and $1 \times 10^{-2}$ for solving CIFAR-10,
    \item Adam \citep{kingmaAdamMethod2015} with a learning rate of $2\times10^{-4}$, and 
    \item Adagrad \citep{duchiAdaptiveSubgradient2011} with a learning rate of $1\times10^{-1}$.
\end{itemize}
We selected the optimizers based on their potential to yield diverse models given the same dataset and model initialization. 
Specifically, to update weights: SGD uses only the gradient; Adam, regarded as the \emph{de~facto} choice for machine learning, uses the gradient's first and second moments; and Adagrad uses only the gradient's second moment.
We do not use weight decay for any of our optimizers. 

\paragraph{Stopping Criteria.} 
We train the MLP models using SGD on MNIST until they reach a fixed loss value of 0.2.
For this architecture and task, no overfitting was detectable and thus we selected a reasonable, fixed loss to terminate on.
In contrast, we train the Conv-2 models with both Adam and Adagrad, 
and the Resnet models with SGD,
on CIFAR-10 until they reach a validation loss minimum (determined retrospectively after 100, 25 and 90 epochs of training respectively).
The crosses in Figure \ref{fig:main-training-loss} indicate the loss minima.

\paragraph{Repetitions and Random Seeds.} 
We use 10 different random seeds for generating model initializations, and 5 different random seeds for partitioning the training set into a train and validation set.

\paragraph{Training Trajectories.}
The training accuracy curves can be found in Figure \ref{fig:main-training-acc} and 
the training loss curves can be found in Figure \ref{fig:main-training-loss}.

\subsection{Pruning Trajectory and Effective Density}\label{sec:method-sparsity}

After training, we use layer-wise magnitude pruning to obtain a pruning trajectory. This process is as follows:
\begin{enumerate}
    \item For each layer, prune the smallest 2\% of weights by magnitude.
    \item Evaluate the pruned subnetwork with a validation set.
    \item Repeat until all weights have been pruned.
\end{enumerate}

Remark that this is \emph{not} iterative magnitude pruning as we do not rewind the model to initialization, or train the model between pruning steps.

Given a threshold representing the acceptable decrease (or increase) in accuracy (or loss), we calculate the \emph{effective density} of a model as the percentage of weights remaining at the first pruning cycle for which the model degrades beyond the acceptable threshold.
In our visualizations we used a threshold of a 5\% decrease in validation accuracy.


\newpage
\section{Experimental Results}\label{sec:main-finding}

In this section we will first show the pruning trajectories, effective densities and absolute prunability of different models. 
Based on the effective densities we reject the null hypothesis.
We discuss the effect from different optimizers and architectures.
Lastly, we investigate the effect of initalizing with Glorot versus He.

\subsection{Pruning Trajectories and Effective Densities}

\begin{figure*}[ht]
    \centering
    \includegraphics[width=\linewidth]{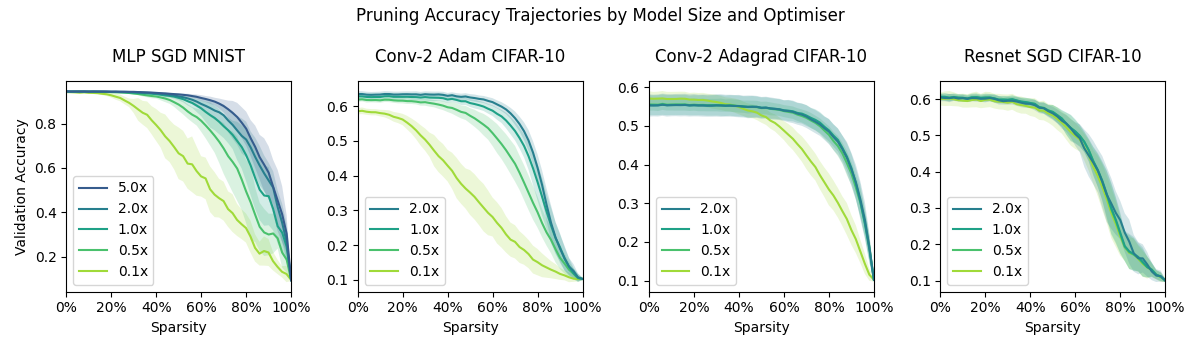}
    \caption{
    The x-axis shows the model sparsities and the y-axis shows the model accuracy. 
    The mean sparsity is plotted as a line and the standard deviation as an opaque area.
    We plot data for models of 5 different architecture sizes (0.1, 0.5, 1, 2, 5) for MLP and of 4 different architecture sizes (0.1, 0.5, 1, 2) elsewhere. }
    \label{fig:main-pruning-trajectories}
    \vspace{1cm}
    \centering
    \includegraphics[width=\linewidth]{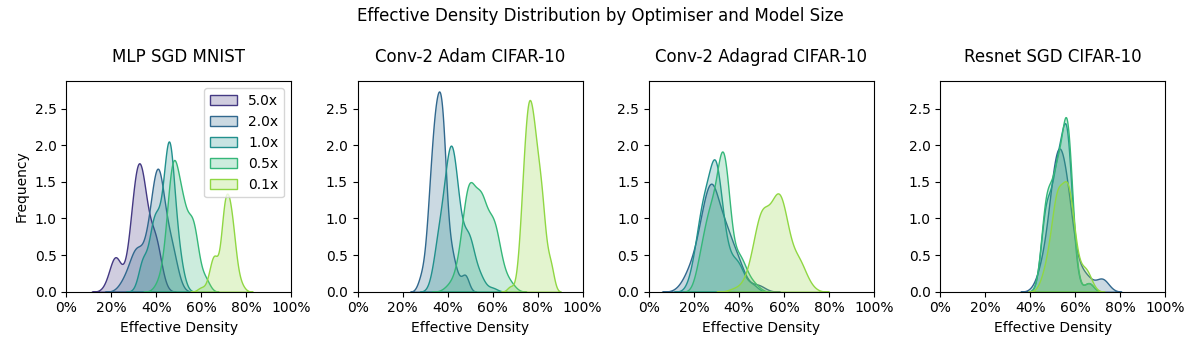}
    \caption{
    The x-axis shows the effective model sparsities and the y-axis shows the frequency by which that effective density was found. 
    We plot data for models of 5 different architecture sizes (0.1, 0.5, 1, 2, 5) for MLP and of 4 different architecture sizes (0.1, 0.5, 1, 2) elsewhere. }
    \label{fig:main-effective-density}
    \vspace{1cm}
    \centering
    \includegraphics[width=\linewidth]{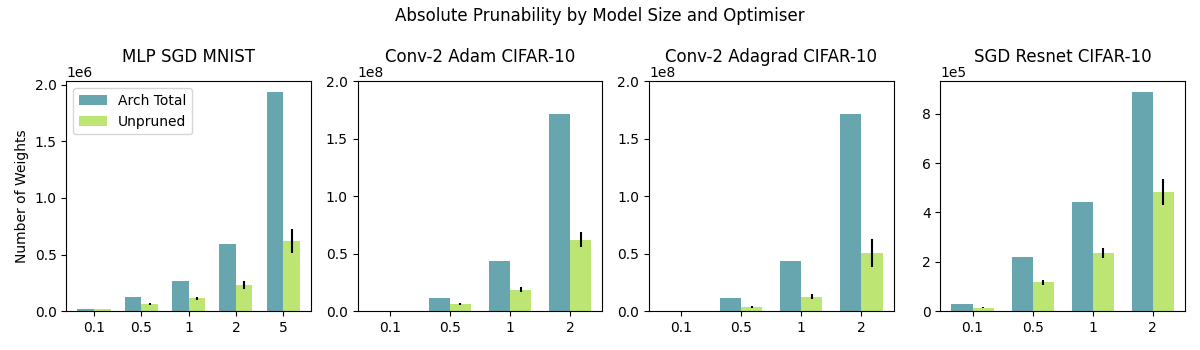}
    \caption{
    For each model size and optimizer we plot the absolute architecture size (arch total) as well as the absolute number of parameters remaining in the pruned network (unpruned). The error bars indicate standard deviations.}
    \label{fig:main-absolute-densities}
\end{figure*}

In Figure \ref{fig:main-pruning-trajectories}, we plot the validation accuracy during the entire pruning trajectory for different model widths.


The effective density distribution is plotted in Figure \ref{fig:main-effective-density}
using the Kernel Density Estimate (KDE) function to estimate the probability distribution (frequency) of the effective densities. 
Apart from for the 0.1x models we find a fair amount of overlap between the effective densities for different model sizes.
This suggests that training regimes may favour a certain amount of effective density.

In Figure \ref{fig:main-absolute-densities} we show how the absolute subnetwork size grows as the model architecture grows.
We find that in absolute terms the pruned models with the biggest architecture size are bigger than unpruned smaller architectures.
Specifically, in absolute terms the \emph{pruned} 5x MLP models are bigger than the \emph{unpruned} smaller architectures.
Similarly, the pruned 2x Conv-2 and Resnet models are bigger than the unpruned smaller models of those architectures.

The smallest networks we train have approximately 24,000 parameters, which is more than two times fewer than the number of data points. 
Even for these heavily underparametrized networks, we find that after training them to a loss of 0.2 we could prune around 30\% of the weights without any substantial impact on performance. 

For overparametrized networks, increasing the width of the layers did not result in substantially more sparsity. 
This means that regardless of the total number of initialized parameters, we could prune around 50\% or more of the weights without causing any substantial harm to the model's performance. 

\subsection{Rejecting the Null Hypothesis}

We calculate the mean effective density for a specific model width by taking the mean of the effective densities of all the models of that width that we trained.
In Table \ref{tab:sizes} we show the mean and standard deviation of the observed effective densities for different model widths. 
For example, for MLP models, we find mean effective densities of 50.7\% and 32.2\% for model widths of 0.5x and 5x respectively.

For each model type (MLP, Conv with Adam, Conv with Adagrad and Resnet) we perform a one-way ANOVA test to verify whether the mean pruned model sizes of different network widths are statistically different. 
We find that for each of the four model types  the $p$-value is less than 0.0001. 
On this basis we reject the null hypothesis that the `core' model size does not change when we vary the network width. 

\begin{table}[hbtp]
\caption{We list the effective densities we would expect to see based on the null hypothesis and a default width of 1x as well as  the mean and the standard deviations of the effective densities (ED) that we observed.
We indicate significance as tested by ANOVA with $\star$ and significance as tested by an unpaired $t$-test by *.}
\begin{center}
\resizebox{\columnwidth}{!}{
\begin{tabular}{|c|c||c|c|c|c|}
\hline
Model &
  Size (x) &
  \begin{tabular}[c]{@{}c@{}}Null Hyp. \\ Mean ED\end{tabular} &
  \begin{tabular}[c]{@{}c@{}}Null Hyp. \\ STD ED\end{tabular} &
  \begin{tabular}[c]{@{}c@{}}Mean \\ ED\end{tabular} &
  \begin{tabular}[c]{@{}c@{}}STD \\ ED\end{tabular} \\ \hline \hline
\multirow{5}{*}{$\text{MLP}^\star$}           & 0.1 & 100        & 0         & $70.6^*$ & 3.6 \\ \cline{2-6} 
    & 0.5 & 91.0       & 0.8       & $50.7^*$ & 4.8 \\ \cline{2-6} 
    & 1   & \emph 43.3 & \emph 4.6 & 43.3     & 4.6 \\ \cline{2-6} 
    & 2   & 19.5       & 2.1       & $39.0^*$ & 5.7 \\ \cline{2-6} 
    & 5   & 6.0        & 0.6       & $32.2^*$ & 5.6 \\ \hline \hline
\multirow{4}{*}{\begin{tabular}[c]{@{}c@{}}Conv\\ ($\text{Adam}^\star$)\end{tabular}} & 0.1 & 100        & 0         & $77.8^*$ & 3.6 \\ \cline{2-6} 
    & 0.5 & 100        & 0         & $54.0^*$ & 5.9 \\ \cline{2-6} 
    & 1   & \emph 43.0 & \emph 5.3 & 43.0     & 5.3 \\ \cline{2-6} 
    & 2   & 11.0       & 1.3       & $36.4^*$ & 3.9 \\ \hline \hline
\multirow{4}{*}{\begin{tabular}[c]{@{}c@{}}Conv\\ ($\text{Adagrad}^\star$)\end{tabular}} & 0.1 & 100        & 0         & $55.8^*$ & 6.7 \\ \cline{2-6} 
    & 0.5 & 96.7       & 6.7       & $32.1^*$ & 5.5 \\ \cline{2-6} 
    & 1   & \emph 29.3 & \emph 5.8 & 29.3     & 5.8 \\ \cline{2-6} 
    & 2   & 7.5        & 1.5       & $29.5^*$ & 6.9 \\ \hline \hline
\multirow{4}{*}{$\text{ResNet}^\star$}         & 0.1 & 100        & 0         & $55.0^*$ & 4.7 \\ \cline{2-6} 
    & 0.5 & 98.6       & 2.8       & $53.1^*$ & 4.4 \\ \cline{2-6} 
    & 1   & \emph 53.1 & \emph 4.4 & 53.1     & 4.4 \\ \cline{2-6} 
    & 2   & 26.4       & 2.3       & $54.4^*$ & 6.0 \\ \cline{1-6}
\end{tabular}
}
\label{tab:sizes}
\end{center}
\end{table}

In Table \ref{tab:sizes} we also calculate the mean and standard deviation of the effective densities that we would expect to see if the null hypothesis were true.
According to the null hypothesis, for the same architecture, training regime and task, the `core' model size should be the same when we decrease or increase the model width. 
Given an architecture, training regime and task, we match the data for every combination of initialization seed and data seed of a wider or more narrow model with the data for a 1x model with the same combination of hyper-parameters.
We then calculate the effective density that we would have seen if the null hypothesis were true as follows.
For a given narrow or wide model we look up what absolute number of weights were unpruned in its 1x counterpart.
We then calculate what proportion that number of weights is of the narrow or wide architecture.
Whenever the number of unpruned weights in the 1x counterpart is larger than the narrow architecture size, 
we say the proportion is 100\% (even if the proportion is technically bigger than 100\%).

For each row in Table \ref{tab:sizes} (other than the 1x rows) we do an unpaired $t$-test and for each row we find that the two-tailed $p$-value is less than 0.0001, which means that our findings are statistically significant.

\subsection{Effect from Different Optimizers and Architectures}

We find that although our MLPs and Resnets are both trained with SGD the pruning curves (Figure \ref{fig:main-pruning-trajectories}) come apart for the MLPs of different sizes, whereas the pruning curves for the Resnets are fairly overlapping.
One thing to note is that multiplying the Resnet layer widths by a factor roughly leads to a parameter increase of that same factor, whereas the effect on parameter number for MLPs is larger.



We trained the same convolutional model architectures with either Adam or Adagrad.
We trained the models until they reached a validation loss minimum. 
However, for Adagrad this resulted in 
all models having similar validation loss at the start of pruning, whereas for Adam, halting training when models reach minimum validation loss did not result in models with the same pre-pruning accuracy.
This difference in training dynamic may help explain why the Adagrad pruning curves overlap more.



\subsection{Glorot Leads to Lower Densities than He}

Glorot (also known as Xavier) initialization involves sampling numbers from the uniform probability distribution over the interval $[- a,a]$
where $ a = \sqrt{\frac{6}{n_{\text{in}} + n_{\text{out}}}}$
with $n_{\text{in}}$ being the number of units in the previous layer (input units) and $n_{\text{out}}$ being the number of units in the next layer (output units).
He (also known as Kaiming) 
initializes weights by drawing numbers from the uniform probability distribution over the interval
 $[-b,b]$ where $b= \sqrt{\frac{6}{n_{\text{in}}}}$. 

\newpage

\clearpage

\begin{figure}[h]
    \centering
    \includegraphics[width=0.8\linewidth]{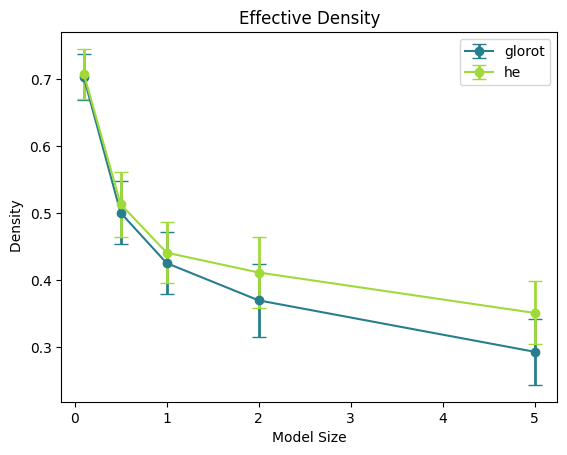}
    \caption{The y-axis shows the model densities and the x-axis shows the model size. We plot data for Glorot initialization in dark green and data for He in light green. The dots show the average effective density and the error bars indicate the standard deviations.}
    \label{fig:avg-glorot-vs-he}
    \vspace{0.4cm}
\end{figure}

For the fully connected and convolutional models and every combination of hyper-parameters we trained one model using Glorot and one model using He.
In Figure \ref{fig:avg-glorot-vs-he} we show the distribution of effective densities that we found for Glorot and He initialization schemes and different model widths.

As the width of the layers increases, the difference in effective density between Glorot and He initializations also increases. 
Networks initialized with the He scheme require fewer epochs to converge and have higher effective density. 

However, the difference in effective density is relatively small between the two initialization schemes, which suggests that the cause for propensity to density does not lie in initialization.

\section{Investigating the Mechanism behind the Implicit Inductive Bias towards Density}\label{sec:hypotheses}

In this section we explore three hypotheses for our finding that wide and narrow models have similar effective density:
\begin{enumerate}
    \item The weights that are small at initialization are the same weights that are small and unimportant after training. Since the proportion of small weights at initialization remains constant (regardless of model width) we see the same effective densities;
    \item The functions that wider models converge on are qualitatively different (and need more parameters to implement);
    \item Given a larger architecture the training regime makes use of this architecture by separating tasks out more.
\end{enumerate}

\vspace{-0.4cm}

\subsection{Comparing Initialization and Trained Weight Magnitudes}\label{sec:init-magnitudes}

We hypothesize that a similar effective density between wide and narrow models is due to wide and narrow models having the same proportion of small weights at initialization, and due to these weights remaining prunable after training.

We looked at the connections that had the lowest magnitude weights at initialization
as well as the connections that had the lowest magnitude weights after training.
More specifically, we looked at the overlap between the smallest 40\% weights at initialization and the smallest 40\% after training.
By random chance we would expect the overlap between two random masks of 40\% to be 16\%.
This is because under the assumption of independence, we can multiply the probabilities as follows $0.4 \cdot 0.4 = 0.16$.



For MLPs in Table \ref{tab:mlp-overlap} we find that the overlap between small initialized weights and small trained weights is larger than chance for model width 0.1x. Surprisingly, we find that the overlap is \emph{smaller} than chance for the first two layers of the overparameterized models.
In contrast, for convolutional models trained with Adam in Table \ref{tab:conv-overlap} we see more overlap than we would expect by chance for \emph{all} layers.

Based on these findings we provisionally conclude that for MLP models our main finding cannot be explained by the initialization.
However, for convolutional models trained with Adam, initialization may play a role in the models' propensity for density.

\begin{table}[h]
\caption{Overlap between the lowest 40\% of weights at initialization and after training by model size and layer (fully connected layer 1, fully connected layer 2, projection) for MLP and Glorot initialization for 10 different initializations.}
\begin{center}
\resizebox{0.51\columnwidth}{!}{
\begin{tabular}{|c||c|c|c|}
\hline
Size (x) & fc1 & fc2 & proj \\ 
\hline
\hline
0.1 & 25.5 & 22.1 & 34.1 \\
\hline
0.5 & 10.8 & 12.1 & 19.3 \\
\hline
1   & 7.2  & 9.3  & 16.7 \\
\hline
2   & 4.9  & 7.1  & 15.9 \\
\hline
5   & 3.1  & 4.6  & 15.9 \\
\hline
\end{tabular}\label{tab:mlp-overlap}
}
\end{center}
\end{table}

\begin{table}[h]
\caption{Overlap between the lowest 40\% of weights at initialization and after training by model size and layer (convolutional layer 1, convolutional layer 2, fully connected layer 1, fully connected layer 2, projection) for Adam and Glorot initialization for 10 different initializations.}
\begin{center}
\resizebox{0.8\columnwidth}{!}{
\begin{tabular}{|c||c|c|c|c|c|}
\hline
Size (x) & conv1 & conv2 & fc1 & fc2 & proj \\ 
\hline
\hline
0.1     & 59.2  & 58.3  & 47.1 & 18.4 & 28.5 \\ \hline
0.5     & 58.6  & 59.3  & 48.3 & 19.5 & 28.8 \\ \hline
1       & 58.7  & 59.6  & 45.0 & 21.8 & 29.3 \\ \hline
2       & 58.9  & 58.8  & 40.1 & 24.7 & 33.4 \\ \hline
\end{tabular}\label{tab:conv-overlap}
}
\end{center}
\end{table}

\subsection{Functional Similarity Between Models of Different Widths}

If wider models were qualitatively different from narrow models,
then this could explain why the implementations of those functions are larger, 
i.e. why in absolute terms wider models use so many unprunable parameters.
In this section we investigate whether wider models are qualitatively different.

\begin{figure}
    \centering
        \includegraphics[width=0.7\linewidth]{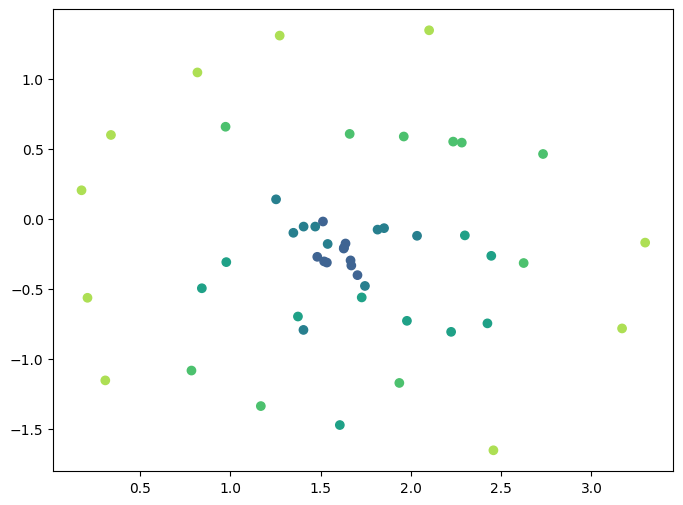}
        \caption{This is a t-SNE plot of the validation correctness vectors. Each dot corresponds to an MLP model trained with SGD. Darker dots correspond to bigger models.}
        \label{fig:sgd-function-tsne}
        \vspace{1cm}
\end{figure}
\begin{figure}
        \centering
        \includegraphics[width=0.7\linewidth]{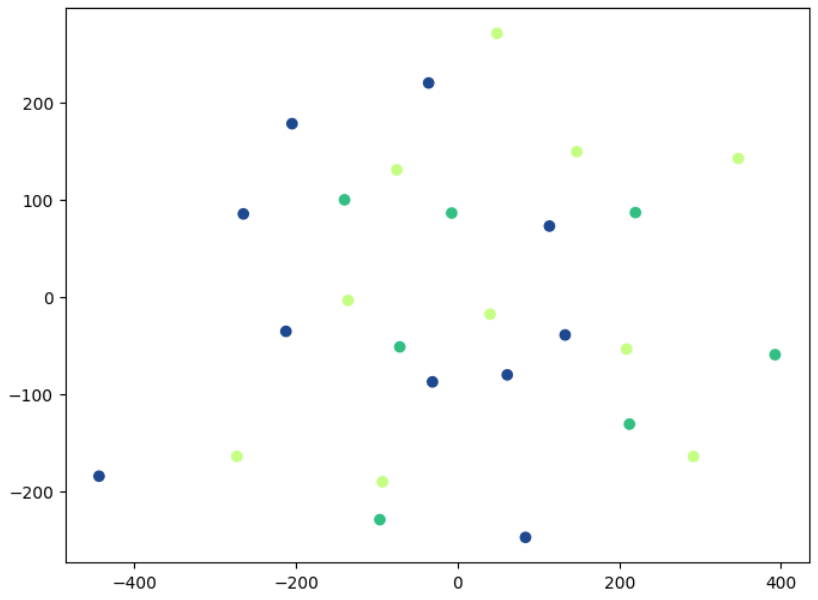}
        \caption{This is a t-SNE plot of the validation correctness vectors. Each dot corresponds to a Conv-2 model trained with Adam. Each color corresponds to a model size.}
        \label{fig:conv-function-tsne}
        \vspace{1cm}
\end{figure}



A crude approximation of model functionality is how the model classifies datapoints in the validation set. 
For each model we create a vector of the length of the validation set and set each entry to 0 or 1 depending on whether that datapoint was classified correctly. 

Some limitations include: 
1) we compress the information by taking the correctness of the classification instead of comparing logits;
2) we only consider similarity on the validation set, which does not give us information about the similarity on data outside of the training distribution.

In Figure \ref{fig:sgd-function-tsne} we plot a t-SNE \citep{vandermaatenVisualizingData2008} of the validation correctness vectors for MLP models and 
in Figure \ref{fig:conv-function-tsne} we plot the t-SNE for Conv models (trained with Adam). 
We plot models for 10 different initialization seeds and one data seed.

For MLP the biggest models form a cluster, whereas the smallest models are very dissimilar.
Note that the validation accuracy at the end of training is very high for all MLP trained models (over 90\%) so the percentage of validation datapoints for which the models can disagree is quite small.
For MLP a difference between functions (between different model sizes) may play a role in the difference in the absolute number of parameters needed for the implementation. 

For Conv, we plot only the smallest three model sizes as we were able to calculate their individual validation accuracies on a standard GPU.
We find that there is no clear pattern distinguishing different model sizes.
For Conv we tentatively conclude that the large difference in `core' model size between networks with different architecture widths is not due to a difference in the implemented functions.

Overall, more investigation is needed to accept or reject this hypothesis.

\subsection{Activation Sparsity and Selectivity}\label{sec:selectivity}

\citet{parkRepresentationsLearnt2022} study the effects of layer width on neuron activations. 
In particular, they study the Hoyer sparsity \citep{hoyerNonnegativeMatrix2004} of activations as well as the class selectivity, also referred to as CCMAS (class-conditional mean activity selectivity) \citep{morcosImportanceSingle2018}.
The Hoyer sparsity is calculated based on the ratio of the L1 norm (sum of absolute values) to the L2 norm (square root of the sum of squares) of a vector.
The class selectivity of a unit is defined as the difference between the mean activation of the class with the largest mean activation, and the average mean of all other classes.

The authors find that activation sparsity and class selectivity for an MLP with a single hidden layer trained on MNIST, are highest when the model is trained with Adagrad, followed by Adam, and finally SGD.
They find that increasing the width from roughly 300 units to 600 units (2x) increases Hoyer sparsity and class selectivity for both Adagrad and Adam and has a mixed effect on SGD.

We relate this to our findings by noticing that an increase in width leads to both more parameters being used in absolute terms (as we saw in Section \ref{sec:main-finding}) and generally leads to more activation sparsity and class selectivity (as shown by \citet{parkRepresentationsLearnt2022}). 
However, for SGD specifically, an increase in width does not necessarily lead to more activation sparsity and class selectivity.

We conclude that more work is needed to confirm or reject the hypothesis that the implicit inductive bias towards density that we find can be explained via a decrease in polysemanticity and superposition \citep{nguyenMultifacetedFeature2016, elhageToyModels2022} as a network increases in size.
We propose two investigations for future work: 
1) an investigation into overlap between circuits found via ACDC \citep{conmyAutomatedCircuit2023} for each class; and
2) an investigation into unit polysemanticity based on a more general interpretation of features than class.

 

\section{Discussion}

In summary, our main results demonstrate that for our feed-forward models and tasks, as the model architecture increases in size, 
the absolute number of unprunable parameters increases, but the proportion of unprunable parameters is relatively stable.


Additionally, we explored three hypotheses for why these results hold and provided preliminary evidence in favour and against them. 
Notably we found different results for MLP models trained with SGD and Conv models trained with Adam.

\paragraph{Limitations and Future Work.}

Further investigations will be required to verify whether our results generalize to other architectures and tasks.

One limitation of our approach to calculating effective density is that we find a subnetwork by magnitude-based pruning. 
However, this method is unlikely to find the optimal subnetwork of a given size since some small weights may be crucial whereas some bigger weights may be unimportant.
We propose to use the edge-popup algorithm \citep{ramanujanWhatHidden2020} to find the optimal subnetwork of a trained network.

Another approach to calculating effective density might be to use a measure of effective dimensionality as used in singular learning theory such as the Real Log Canonical Threshold (RLCT) \citep{lauQuantifyingDegeneracy2023}
to capture a model's `true' size as the width is scaled.





\section*{Acknowledgements}
This work was supported by UK Research and Innovation [grant number EP/S023356/1], in the UKRI Centre for Doctoral Training in Safe and Trusted Artificial Intelligence (\url{www.safeandtrustedai.org}); and the DAAD Research Internships in Science and Engineering (RISE Worldwide) scheme.




\bibliography{references}



\end{document}